\documentclass[manuscript,screen,nonacm]{acmart}  

\setcopyright{none}  
\authorsaddresses{}  

\acmDOI{}  

\usepackage[utf8]{inputenc}
\usepackage{natbib}
\usepackage{graphicx}
\usepackage{amsmath}
\usepackage{amsfonts}
\usepackage{hyperref}
\usepackage{booktabs}
\usepackage{array}
\usepackage{tabularx}
\usepackage{xcolor}
\usepackage{float}
\usepackage{cleveref}
\usepackage[normalem]{ulem}
\useunder{\uline}{\ul}{}
\usepackage{colortbl}

\usepackage{subcaption}
\usepackage{enumitem}

\title{Internal Deployment Gaps in AI Regulation}

\author{Joe Kwon}
\authornote{Work done as a Fall Fellow at GovAI. Correspondence to \texttt{joekwon333@gmail.com} and \texttt{scasper@mit.edu}}

\author{Stephen Casper}
\affiliation{%
  \institution{MIT CSAIL}
  \city{Cambridge}
  \state{Massachusetts}
  \country{USA}
}

\date{}

\begin{document}
\raggedbottom

\begin{abstract}
Frontier AI regulations primarily focus on systems deployed to external users, where deployment is more visible and subject to outside scrutiny. However, high-stakes applications can occur internally when companies deploy highly capable systems within their own organizations, such as for automating R\&D, accelerating critical business processes, and handling sensitive proprietary data. This paper examines how frontier AI regulations in the United States and European Union in 2025 handle internal deployment. We identify three gaps that could cause internally-deployed systems to evade intended oversight: (1) scope ambiguity that allows internal systems to evade regulatory obligations, (2) point-in-time compliance assessments that fail to capture the continuous evolution of internal systems, and (3) information asymmetries that subvert regulatory awareness and oversight. We then analyze why these gaps persist, examining tensions around measurability, incentives, and information access. Finally, we map potential approaches to address them and their associated tradeoffs. By understanding these patterns, we hope that policy choices around internally deployed AI systems can be made deliberately rather than incidentally.
\end{abstract}

\maketitle


\section{Introduction} \label{sec:intro}

Consider scenarios emerging in frontier AI development in which companies deploy advanced systems internally rather than releasing them to customers. 
For example, AI development tools like Claude Code can be configured to access proprietary codebases and infrastructure within a company like Anthropic to accelerate internal software development \citep{anthropic2025claudecode}.
Anthropic CEO, Dario Amodei, recently reported that approximately 90\% of internal code was AI-generated \citep{amodei2025dreamforce}.
Meanwhile, Sam Altman recently announced OpenAI's goal to create a ``true automated AI researcher by March of 2028'' \citep{altman2025tweet}, which could allow it to automate portions of AI research: analyzing experiments, generating code, and speeding up the development of next-generation models \citep{rand2024aiautomation}.
Companies may also use frontier AI systems to moderate content, analyze private user behavior data, or optimize engagement algorithms with access to detailed personal information. Elon Musk recently wrote that ``Grok will literally read every post and watch every video (100M+ per day) to match users with content they’re most likely to find interesting.''\citep{musk2025tweet} 
In each case, these systems operate with privileged access, handle sensitive data, and perform high-stakes tasks. Yet these applications all happen with entirely selective visibility to external observers, competitors, and regulators.

This raises a question for policymakers: \textit{When do regulatory obligations apply?} If obligations trigger upon ``deployment'' or when systems are ``placed on the market,'' is an AI system deployed only internally subject to regulatory requirements? External deployment is far more visible to the public and lawmakers and generates natural oversight signals. Yet high-stakes applications can still occur entirely internally in environments with privileged infrastructure access, reduced safeguards, and application to complex operational tasks \citep{stix2025aibehindcloseddoors}.

This paper examines a pattern in frontier AI regulation\footnote{As of January 2026: California's SB 53 was signed into law on September 29, 2025. New York's RAISE Act was signed into law in December 2025. The US/federal AIREA was introduced on September 29, 2025. Finally, the EU AI Act's general-purpose AI obligations took effect on August 2, 2025, with an approved Code of Practice published July 10, 2025.}: existing regulatory frameworks targeting frontier AI systems contain gaps that may limit oversight of internal deployment \citep{acharya2025internalrisks}. Specifically, we analyze the European Union's AI Act (EU AIA) and General-Purpose AI Code of Practice (GPAI COP) \citep{euaiact2024},\footnote{For the European context, our analysis incorporates the General Purpose AI Code of Practice along with the AI Act itself. Because the Code defines how and when regulatory obligations trigger (specifying the ``Measures'' that providers must implement for compliance with Articles 53 and 55) it is decisive for signatories in determining whether internal models face oversight or exemption.} California Transparency in Frontier Artificial Intelligence Act (SB 53) \citep{casb532025}, New York Responsible AI Safety and Education (RAISE) Act \citep{nyraiseact2025}, and the US' Federal Artificial Intelligence Risk Evaluation Act (AIREA) \citep{hawley2025airiskevaluation}---frontier AI regulations enacted or proposed in 2025. We identify three gaps: 

\begin{enumerate}[nosep]
    \item \textbf{Scope ambiguity that allows internal systems to evade regulatory obligations:} Regulatory frameworks vary widely in whether they cover internal deployment. Some focus exclusively on external deployment, while others include broad research and development (R\&D) exemptions with unclear boundaries between exempt research and operational internal use. Even when internal deployment is explicitly covered, questions remain about what modifications trigger re-assessment or how to classify systems deployed in certain jurisdictions.
    \item \textbf{Static compliance criteria for continuously evolving internal systems:} Internal systems may evolve more rapidly than externally deployed ones because they lack external market pressures for discrete release cycles and version stability. Without scheduled releases or user-facing versions, internal systems can be continuously fine-tuned on proprietary data and integrated with other internal systems, with changes tending to happen more continuously rather than at discrete checkpoints.
    \item \textbf{Information asymmetries that subvert regulatory awareness and oversight:} While regulators face visibility challenges with almost all AI systems, internal deployment creates far deeper information asymmetries. External deployment generates observable signals while internal deployment does not. Regulators cannot independently reliably determine what internal systems exist, what capabilities they possess, or how they are used.
\end{enumerate}

\definecolor{headercolor}{HTML}{E2E3E5} 
\definecolor{altrowcolor}{HTML}{F8F8F8} 
\begin{table}[!t]
\centering
\caption{\textbf{How the Three Structural Gaps Manifest Across Regulatory Frameworks.} All four frameworks exhibit gaps that may allow internal deployment to escape intended oversight. \textit{Scope ambiguity}: a lack of clarity on whether internal deployment triggers regulatory obligations. \textit{Static compliance}: assessment criteria that cannot keep pace with continuously evolving systems. \textit{Information asymmetry}: the inability of regulators to independently determine what internal systems exist or how they operate.}
\label{tab:frameworks}
\resizebox{\textwidth}{!}{%
\begin{tabular}{@{}p{2.6cm}p{4.6cm}p{4.6cm}p{4.6cm}@{}}
\toprule
\rowcolor{headercolor}
\textbf{Framework} & \textbf{Scope Ambiguity} & \textbf{Static Compliance} & \textbf{Information Asymmetry} \\
\midrule
EU AI Act + GPAI Code of Practice & Provisions suggest internal deployment triggers coverage when AI systems are "put into service" for "own use." The main exception is narrow: Article 2(6) exempts only systems specifically developed and deployed for the sole purpose of scientific R\&D. GPAI models trigger coverage upon integration into AI systems. & For GPAI models with systemic risk, reassessment required when capabilities change materially, but providers determine what constitutes a "material" change to their safety justification & Detailed Model Reports required when GPAI models are placed on the market; for internal systems, primary documentation is the high-level Safety and Security Framework. Serious Incident Reporting applies across the model lifecycle. \\
\addlinespace
\rowcolor{altrowcolor}
California SB 53 & Explicitly addresses internal use, but key trigger terms like 'extensively' are undefined; no registration mechanism to identify covered developers & Quarterly internal use summaries and annual framework self-review; no definition of what system changes require updated disclosures between reporting periods & Requires confidential risk summaries, not underlying evaluation data; 15-day incident reporting; whistleblower protections for covered employees; no independent verification mechanism \\
\addlinespace
NY RAISE Act & Explicitly addresses internal use, but key trigger terms like 'extensively' are undefined; registration requirement helps identify covered developers; geographic scope limited to New York & Quarterly internal use summaries and annual framework self-review; no definition of what system changes require updated disclosures between reporting periods & Requires confidential risk summaries, not underlying evaluation data; 72-hour incident reporting; no whistleblower protections; registration identifies developers but not their internal systems; no independent verification mechanism \\
\addlinespace
\rowcolor{altrowcolor}
US Federal AIREA & Regulatory enforcement only applies when systems are released externally; systems used only within the company have no enforcement mechanism regardless of their capabilities or risks & Annual updates to Congress; framework assumes deployment is a discrete one-time event rather than a gradual transition & Regulators can request code, data, and model weights, but must already know which companies to investigate; purely internal systems generate no signals \\
\bottomrule
\end{tabular}%
}
\end{table}

These gaps reflect inherited assumptions from product-market regulation: that development and deployment are temporally distinct, that risks emerge from external use, and that regulatory obligations attach at market entry. For AI systems applied to high-stakes internal operations, these assumptions create blind spots. Given these challenges, this paper makes two contributions: 

\begin{enumerate}[nosep]
    \item We observe three specific regulatory gaps (scope ambiguity, static compliance criteria, and information asymmetries) that limit oversight of internal deployment across high-stakes use cases.
    \item We analyze whether these gaps appear across the EU AIA and related GPAI COP, California's SB 53, New York's RAISE Act, and the US AIREA proposal, given different jurisdictional contexts and design processes.
\end{enumerate}

\section{Why Internal Deployment Matters} \label{sec:why}

Internal deployment is the act of making an AI system available for access and usage exclusively within the developing organization. This distinguishes it from external deployment (making systems available to customers or the public) and from development where tools are built but not yet operational. Internal deployment means the system is functioning, performing tasks, and generating value, but only within the organization that created it.

\subsection{Distinctive Characteristics} \label{sec:characteristics}

Internal deployment of frontier AI systems differs from external deployment along three dimensions: system configuration, access patterns, and applications.

\textbf{Internally deployed systems often operate with reduced safeguards and expanded capabilities.} Systems deployed externally are typically deployed with substantial safety measures: fine-tuning to refuse harmful requests, content filters, and usage restrictions \citep{databricks2024guardrails}. Internally deployed systems are subject to fewer restrictions. The company controls the environment, and users are employees rather than anonymous actors, enabling acceptance of risks from reduced safeguards while gaining performance benefits. For example, frontier AI companies often use ``helpful-only'' versions of models internally---systems designed to comply with any request---for research and development where unrestricted models provide more value \citep{openai2025safetyeval}. Internal systems can also receive expanded affordances never granted externally: read access to proprietary codebases, write access to experimental code, the ability to launch training runs, and integration with other internal AI systems. An external chatbot cannot execute code on company servers or modify production systems, but an internal AI research system might have all these capabilities. This distinction in configuration means that the risk profile can differ substantially from external counterparts built from the same underlying model.

\textbf{Internal deployment can permit deeper configuration access and system orchestration.} External deployment involves tiered, metered, and logged access through defined interfaces with limited permissions. Internal deployment in some cases permits configuration-level access: internal users may modify system prompts, adjust behavioral guidelines, or change integration with other tools. An employee with appropriate permissions could redirect an internal AI system toward harmful goals, either deliberately (as an insider threat) or inadvertently (via poorly-designed prompts). Internal access can also mean AI systems accessing other AI systems via orchestration layers, where one system decides which others to invoke, or multi-agent configurations where multiple systems collaborate with minimal human oversight. These configurations can produce emergent behaviors that would not appear in any individual system operating alone.

\textbf{Internal deployment enables higher-stakes applications.} External deployments serve customers at arm's length with limited access to sensitive data or infrastructure. Internal deployment is different. A company might deploy a highly capable system with full access to detailed user behavioral data, analyzing interaction patterns, emotional responses, and social connections to optimize engagement algorithms. This capability would raise significant concerns if externally accessible, but generates enormous value when applied internally. Companies might similarly deploy advanced systems to automate portions of scientific research, optimize complex business strategies using proprietary intelligence, or accelerate AI research by designing experiments, analyzing results, and writing code. Multiple frontier AI companies are pursuing this last application: Anthropic CEO Dario Amodei has stated that Claude is ``playing this very active role in designing the next Claude'' \citep{axios2025amodei}, while OpenAI CEO Sam Altman has set internal goals of having ``an automated AI research intern by September of 2026'' and ``a true automated AI researcher by March of 2028'' \citep{altman2025tweet}. The economic incentives are powerful across these applications. Automating scarce expertise, whether in user psychology, strategic planning, or AI development, provides competitive advantages too valuable to forego. In competitive landscapes where capability thresholds provide decisive advantages, companies have strong incentives to pursue internal deployment of their most capable systems for applications that are too strategically important or risky to expose externally.

\subsection{Governance Implications} \label{sec:governance}

These distinctive characteristics of internal deployment (differential configuration, privileged access, and high-stakes applications) create challenges for effective governance.

\textbf{Effective governance requires basic visibility into what actors are doing, what capabilities exist, and what risks may be materializing.} Despite the essential role of domain awareness for crafting sound and effective policy, governments and the public generally lack fundamental visibility into internal AI deployment \citep{bommasani2025advancing, casper2025pitfalls}. They cannot reliably determine which systems are deployed internally, what capabilities those systems possess, how they are being used, whether stated safety protocols are followed, or what concerning behaviors emerge during internal operation. For externally deployed systems, multiple information sources provide visibility: market entry, user reports, researcher analysis, media coverage, and competitive scrutiny. Internal deployment generates no such signals. The information flows unidirectionally from companies outward and only when companies choose to disclose. This information asymmetry stands in contrast to regulatory approaches in other high-risk domains. Nuclear facilities have NRC inspectors on-site with real-time access to monitoring data \citep{nrc2020oversight}. Pharmaceutical companies must submit detailed trial protocols and results to FDA before and during testing \citep{fda2015indprotocols}. Here, regulatory visibility begins before external deployment, during phases when risks are least well-characterized. Current AI regulation operates on the opposite principle: minimal mandatory disclosure for internal development and use, with obligations attaching primarily when systems are released externally.

\textbf{Traditional oversight mechanisms assume systems operate within well-defined boundaries that can be monitored and constrained.} Internal deployment challenges this assumption. When an AI system can invoke other AI systems, modify code those systems depend on, and launch training runs that alter future system behavior, no single checkpoint (e.g., input filtering, output review, periodic audit) captures the compound risk. The system is not a discrete product but a node in an evolving network of interdependencies. This poses a fundamental problem: conventional regulatory approaches are designed for standalone products with stable boundaries, but internal deployment creates configurations where the very features that generate value (e.g., deep integration, extensive permissions, multi-agent orchestration) make the ``system'' difficult to even delineate, let alone assess. The result is a governance gap where the systems posing the most complex control challenges operate in contexts least accessible to standard oversight mechanisms.

\textbf{Internal deployment enables private actors to make consequential decisions about AI capabilities and applications without external accountability.} For external deployments, public scrutiny provides at least some accountability; when problems become visible, companies often respond with meaningful improvements. Internal deployment eliminates this mechanism entirely. The public should care what happens behind corporate walls even if externally released systems are safe because internal deployment is not parallel to external deployment but upstream of it. When companies use their most capable systems to accelerate AI research, optimize training approaches, or automate capability development, the pace and direction of progress on external systems is shaped by internal choices no outside observer can scrutinize. A company's internal deployment decisions today determine what external systems exist tomorrow, and on what timeline.

\begin{table}[!t]
\centering
\caption{\textbf{How Internal Deployment Differs from External Deployment.} Internal deployment of frontier AI systems differs systematically from external deployment across three dimensions (Section 2.1), creating distinct risk profiles and governance challenges (Section 2.2).}
\label{tab:internal-external}
\resizebox{\textwidth}{!}{%
\begin{tabular}{@{}p{2.8cm}p{6cm}p{6cm}@{}}
\toprule
\rowcolor{headercolor}
\textbf{Dimension} & \textbf{External Deployment} & \textbf{Internal Deployment} \\
\midrule
\textbf{System Configuration} (\S2.1) & Substantial safety measures: refusal training, content filters, usage restrictions; externally-facing guardrails & Reduced safeguards; ``helpful-only'' models designed to comply with any request; expanded affordances including code execution, training run launches, write access to experimental code \\
\addlinespace
\rowcolor{altrowcolor}
\textbf{Access Patterns} (\S2.1) & Tiered, metered, and logged access through defined APIs with limited permissions; users are anonymous actors & Configuration-level access: modify system prompts, adjust behavioral guidelines, change tool integrations; multi-agent orchestration where systems invoke other systems with minimal human oversight \\
\addlinespace
\textbf{Applications} (\S2.1) & Customer-facing services at arm's length; limited access to sensitive data or proprietary infrastructure & High-stakes applications: user behavior analysis with full data access, AI R\&D automation (``designing the next Claude''), strategic optimization with proprietary intelligence \\
\midrule
\rowcolor{altrowcolor}
\textbf{Visibility to Regulators} (\S2.2) & Observable market signals: user reports, researcher analysis, media coverage, competitive scrutiny; multiple information sources regardless of company disclosure & No external signals generated; information flows unidirectionally from companies outward; regulators cannot independently determine what systems exist, what capabilities they possess, or how they are used \\
\addlinespace
\textbf{System Boundaries} (\S2.2) & Discrete products with stable boundaries; version stability pressures from customer compatibility requirements & Systems as nodes in evolving networks: invoke other AI systems, modify code, launch training runs; boundaries difficult to delineate; continuous fine-tuning, integration, and forking without natural checkpoints \\
\addlinespace
\rowcolor{altrowcolor}
\textbf{Accountability} (\S2.2) & Public scrutiny provides accountability mechanism; when problems become visible, companies often respond with improvements & No external accountability; internal choices shape pace and direction of progress on \textit{external} systems; today's internal deployment decisions determine tomorrow's external capabilities \\
\bottomrule
\end{tabular}%
}
\end{table}

\section{Three Structural Gaps} \label{sec:gaps}

Current frontier AI regulations are meant to target the types of risks that can emerge from internal deployment, yet their coverage of internal deployment remains incomplete. For example, AIREA directs evaluations to assess ``scheming behavior,'' ``loss-of-control scenarios,'' and AI systems' potential to ``exceed human oversight or operational control''. California's SB 53 targets ``loss of human control'' over frontier systems. These risk profiles describe the risk factors common in internal deployments with unusual precision: systems operating with reduced safeguards, expanded permissions, and minimal external visibility. Despite targeting these risks, regulations provide limited mechanisms for addressing the contexts where they are most likely to manifest. We identify three structural gaps that limit oversight of internal deployment:

\begin{enumerate}
    \item Scope ambiguities that blur the boundary between exempt ``research'' and regulated ``deployment,'' allowing operational internal systems to evade obligations entirely.
    \item Static compliance criteria designed for discrete product releases, which fail to govern systems that evolve continuously through use, modification, and integration.
    \item Information asymmetries that prevent regulators from independently determining what internal systems exist, what capabilities they possess, or how they are being used, which creates a circular problem where oversight requires information that only oversight could compel.
\end{enumerate}

\subsection{Scope and Coverage Gaps} \label{sec:scope}

Scope ambiguity arises when regulations fail to clearly specify whether internal deployment triggers obligations. Across the frameworks analyzed, this ambiguity manifests through three mechanisms: R\&D exemptions with unclear boundaries, deployment definitions that may exclude internal use, and self-defined compliance scope that permits companies to set their own thresholds.

\textbf{The EU AI Act's scope provisions offer multiple pathways for covering internal deployment, though key boundaries remain untested.} Article 3(11) defines "putting into service" to include supply of an AI system "for own use," which can be interpreted to encompass internal deployment. When read alongside Articles 2(1)(a)-(c), this suggests that deploying an AI system internally in the EU triggers the Act's obligations \citep{pistillo2025internal}. Article 2(8), which excludes "research, testing or development activity... prior to being placed on the market or put into service," may not create an exception for internal deployment—rather as it describes activities that precede any deployment, whether internal or external. The narrower exception is Article 2(6), which excludes AI systems "specifically developed and put into service for the sole purpose of scientific research and development." However, this exemption is intended to apply to systems built for scientific R\&D, such as model organisms for safety research or mechanistic interpretability \citep{pistillo2025internal}. Using AI systems to accelerate commercial model development should not qualify, as this constitutes product-oriented research under the Act's framework.

Safety and Security Measure 1.2 requires signatories to assess and mitigate systemic risks "along the entire model lifecycle," including development phases before market placement. The Code of Practice applies only to GPAI models already within the Act's scope; a question is whether internally deployed GPAI models fall within scope as Article 2(1)(a) references GPAI models only in the context of "placing on the market," not "putting into service." But Recital 97 notes that when a provider integrates a GPAI model into an AI system that is put into service, the model "should be considered to be placed on the market." Since internal deployment typically requires system integration (adding affordances, tool access, and infrastructure connections), most operationally deployed internal GPAI models would trigger coverage through this pathway \citep{pistillo2025internal}.

\textbf{For GPAI models with systemic risk, implementation details remain provider-driven within the Act's existing scope.} For GPAI models with systemic risk that already fall within the Act's scope, the Code of Practice grants signatories discretion over implementation details. Specifically, Safety and Security Measure 4.1 permits signatories to "define appropriate systemic risk tiers" and "systemic risk acceptance criteria" within their Safety and Security Frameworks. This self-definition operates only within boundaries already established by the Act and it does not permit providers to determine whether their models fall within scope in the first place. However, it does create latitude in how stringently providers apply mitigations to covered internal systems, potentially allowing developers to characterize internal environments as lower-risk contexts warranting less intensive safeguards.

\textbf{Deployment definitions may independently exclude internal use from coverage.} Even where R\&D exemptions do not apply, deployment definitions may independently exclude internal systems. AIREA illustrates this tension. Section 3(7) defines ``covered advanced artificial intelligence system developer'' broadly as anyone developing advanced AI systems ``for use in interstate or foreign commerce'', while Section 3(8) defines ``deploy'' narrowly as releasing or providing access ``outside the custody of the developer.'' Section 4(a) imposes participation obligations on all covered developers regardless of deployment status, requiring them to provide materials including code, training data, and model weights ``on request.'' A company developing a $10^{26}$ FLOP system for internal research automation would arguably be legally obligated to participate if such R\&D constitutes use ``in interstate or foreign commerce.'' Yet the narrow deployment definition means Section 4(b)'s prohibition on non-compliant deployment provides no enforcement mechanism for systems that never leave developer custody.

\textbf{US state frameworks explicitly address internal use but leave key boundaries undefined.} California's SB 53 and New York's RAISE Act represent the most explicit attempts to address internal deployment among the frameworks analyzed, with nearly identical provisions. Both define ``deploy'' as making a frontier model available to a third party, meaning core transparency report obligations attach to external deployment events. However, both separately require large frontier developers to address internal use through multiple mechanisms: frontier AI frameworks must describe how developers assess catastrophic risk ``resulting from the internal use of its frontier models, including risks resulting from a frontier model circumventing oversight mechanisms'' (SB 53 Section 22757.12(a)(10); RAISE Section 1421(a)(10)); assessments must be reviewed ``as part of the decision to deploy a frontier model or use it extensively internally'' (SB 53 Section 22757.12(a)(4); RAISE Section 1421(a)(4)); incident reports must note whether incidents were associated with internal use (SB 53 Section 22757.13(a)(4); RAISE Section 1422(a)(4)); and large frontier developers must submit confidential quarterly summaries of internal use risk assessments to regulators (SB 53 Section 22757.12(d); RAISE Section 1422(b)(1)).

Yet both statutes share the same critical ambiguity: neither defines what constitutes ``extensive'' internal use, leaving unclear when ongoing development crosses into regulated operational deployment. The term appears without elaboration in the statutory text, providing no guidance on whether thresholds involve duration or volume of use, amount of integration, reliance on outputs, etc. Additionally, while internal use receives confidential regulatory oversight under both frameworks, public transparency reports remain tied to external deployment. The RAISE Act also contains significant geographic limitations as it restricts the scope to frontier models ``developed, deployed, or operating in whole or in part in New York state''.

\textbf{Common challenges: gaps through which internal deployment may escape coverage.} Each regulation analyzed contains mechanisms through which internal deployment may escape intended coverage: broad R\&D exemptions, narrow deployment definitions, self-determined risk thresholds, or combinations of all three. Even those which most directly address internal use, leave key boundaries undefined—most notably the threshold for ``extensive'' internal use that triggers regulatory obligations. The common thread is that regulations designed around a clear develop-then-deploy sequence struggle to classify systems that blur this boundary regarding performing operational work and simultaneously contributing to ongoing development.

\subsection{Static Compliance Criteria for Continuously Evolving Systems} \label{sec:static}

Regulatory compliance typically assumes products with stable characteristics across discrete versions: assess conformity at a defined point, certify, review periodically, and respond if problems emerge. Internal AI deployment challenges this model. Without external pressures for version stability in customer compatibility requirements and public release cycles, internal systems can evolve continuously through fine-tuning, integration, and replication in ways that outpace fixed assessment intervals.

\textbf{Internal systems can evolve continuously in ways that external product releases cannot.} The dynamics of internal deployment differ fundamentally from external product releases. Systems may be fine-tuned regularly on proprietary data, accumulating familiarity with internal infrastructure and processes. A model evaluated at initial deployment could subsequently be trained on the company's codebase, substantially increasing its ability to navigate internal environments. Changes accumulate between assessment periods through gradual adjustments that individually appear minor but collectively shift capabilities significantly. Systems initially used in isolation may be integrated into multi-agent workflows where multiple models collaborate, producing emergent behaviors absent from any individual system. Internal systems can also fork: a single assessed system may spawn multiple variants adapted for different purposes, each evolving along its own trajectory. Where external deployment requires maintaining a canonical product, internal deployment permits and encourages proliferation without natural checkpoints.

\textbf{For GPAI models with systemic risk, the EU framework ties reassessment to capability changes, but materiality determinations remain provider-driven.} For GPAI models with systemic risk, the Code of Practice moves beyond fixed assessment intervals; Safety and Security Measure 1.2 mandates evaluations "along the entire model lifecycle" when specific thresholds are crossed: changes in training compute, inference compute, or affordances. Safety and Security Measure 7.6 requires Model Report updates when capabilities change materially, citing "access to additional tools" or "increase in inference compute" as triggering events. This event-driven approach ties assessment to actual capability changes rather than calendar dates, but introduces interpretive discretion. For example, reassessment obligations activate when providers have "reasonable grounds to believe" their previous safety justification has been "materially undermined"—a standard that relies on provider judgment. For internally deployed systems undergoing continuous fine-tuning, expanded tool access, or integration with other internal systems, cumulative capability gains may not trigger reassessment if each incremental change is characterized as immaterial in isolation. The framework addresses continuous evolution more directly than fixed-interval approaches, but verification depends on provider self-reporting rather than independent monitoring.

\textbf{US frameworks combine fixed intervals with company-defined triggers, but neither mechanism fully addresses continuous evolution.} US frameworks rely primarily on pre-deployment gates and scheduled reassessment. California's SB 53 and New York's RAISE Act both mandate quarterly internal use risk assessment reporting, making them the most frequent among the regulations analyzed. The federal AIREA proposal contemplates ongoing evaluation through annual plan updates to Congress.

Both SB 53 and RAISE require large frontier developers to publish frontier AI frameworks on their websites, conduct annual self-reviews, and publish updates within 30 days of material modifications to the framework (SB 53 Section 22757.12(a)-(b); RAISE Section 1421(a)-(b)). Both also require the framework to describe ``any criteria that trigger updates and how the large frontier developer determines when its frontier models are substantially modified enough to require disclosures'' (SB 53 Section 22757.12(a)(6); RAISE Section 1421(a)(6)). This creates a hybrid approach: fixed quarterly reporting intervals supplemented by company-defined triggers for what model changes warrant disclosure.

However, this hybrid approach shares the EU framework's core limitation: the company determines what counts as a material change. Consider a system assessed in January that undergoes daily fine-tuning, tool integrations, and workflow changes throughout February. By March, accumulated modifications may have substantially altered its capabilities, yet no disclosure obligation triggered if the company's self-defined criteria characterized each incremental change as immaterial. The quarterly checkpoint eventually arrives, but the requirement covers summaries rather than underlying evaluation data. For systems evolving through continuous iteration rather than discrete releases, quarterly intervals may function less as oversight mechanisms than as periodic snapshots of moving targets.

\textbf{Pre-deployment assessment works when deployment is a discrete event, but internal systems may transition gradually without clear checkpoints.} AIREA frames core obligations around pre-deployment assessment with conformity verified before market release. Both SB 53 and RAISE similarly require transparency reports before deploying new frontier models (SB 53 Section 22757.12(c); RAISE Section 1421(c)), though their emphasis on ongoing quarterly reporting provides some visibility into internal operations beyond the pre-deployment gate. This approach works when deployment is a discrete event creating a checkpoint. For internal systems, no such event may occur. A model may transition gradually from experimental use to operational reliance without a moment that clearly constitutes ``deployment.'' The pre-deployment paradigm assumes a gate to pass through; internal deployment may instead involve continuous expansion of use cases, permissions, and integration depth with no single triggering moment.

\textbf{Common challenges: difficulties rooted in internal deployment's inversion of the normal relationship between assessment and evolution.} Regulations employ two primary strategies for governing system evolution: event-driven triggers tied to capability changes (EU, and to a lesser extent the US state frameworks' company-defined criteria) and fixed periodic reviews (all US frameworks). Each addresses the static compliance problem but creates new gaps. Event-driven approaches rely on self-assessed materiality thresholds that providers control. Fixed intervals create predictable oversight but cannot keep pace with systems optimized for continuous iteration. Pre-deployment gates assume discrete transition points that may not exist for internal systems. The underlying challenge is that internal deployment inverts the normal relationship between assessment and evolution: rather than assessing a stable product periodically, regulators must somehow govern systems that evolve through their operational use, where the very activity generating value also generates capability drift.

\subsection{Information Asymmetry and Verification Challenges} \label{sec:asymmetry}

Information asymmetry between regulators and AI developers exists for all AI systems, but internal deployment creates a structural problem distinct in kind: regulators cannot independently determine what internal systems exist, what capabilities they possess, or how they are being used. For externally deployed systems, multiple signals flow to regulators regardless of company disclosure via user reports, researcher analysis, market entry, and competitive scrutiny. Internal deployment generates none of these signals. Information flows unidirectionally from companies outward, and only when companies choose to disclose.

\textbf{Regulators lack authority to obtain the information required to establish jurisdiction.} This asymmetry creates a sequential dependency that undermines enforcement. Normal regulatory oversight follows a clear sequence: determine whether an activity falls within jurisdiction, compel disclosure, assess compliance, enforce requirements. For internal deployment, this sequence breaks down. The information required to establish jurisdiction (training compute, capabilities, use cases, deployment status) is the very information regulators lack authority to compel until jurisdiction is established.

AIREA illustrates this circularity acutely. Section 4(a) creates participation obligations for all covered developers, requiring them to provide materials including code, training data, and model weights ``on request.'' But regulators cannot request materials from entities they do not know are covered. The statute's information-gathering authority presupposes knowledge of which entities to request information from, which is the very knowledge unavailable for internal-only systems. Section 4(b)'s deployment prohibition provides an external enforcement mechanism, but only when systems leave developer custody. A company could operate a highly capable internal system for years, technically in violation of Section 4(a)'s participation requirement, with no mechanism for detection absent whistleblower disclosure or eventual external release.

New York's RAISE Act partially addresses this jurisdictional knowledge problem through a registration mechanism. Section 1428 requires large frontier developers to file disclosure statements with the Department of Financial Services Office before developing, deploying, or operating a frontier model in New York, with renewal every two years. This creates an affirmative obligation for covered entities to identify themselves, providing regulators with a baseline registry of developers subject to oversight. California's SB 53 lacks an equivalent registration requirement, relying instead on the assumption that regulators will independently identify covered developers.

\textbf{Reactive disclosure mechanisms capture failures but not the ongoing operations that precede them.} Regulations attempt to bridge information gaps through mechanisms that surface problems after they occur. The EU's GPAI Code of Practice mandates Serious Incident Reporting (Safety Commitment 9) along the ``entire model lifecycle,'' ensuring regulators are notified of ``serious cybersecurity breaches'' or ``critical infrastructure disruption'' even if a model is never released. Both SB 53 and RAISE require incident reporting, though with different timelines: RAISE mandates 72-hour disclosure of critical safety incidents (Section 1422(c)(1)), while SB 53 allows 15 days (Section 22757.13(c)(1)). Both require expedited 24-hour reporting when incidents pose imminent risk of death or serious physical injury (SB 53 Section 22757.13(c)(2); RAISE Section 1422(c)(2)).

SB 53 additionally requires large frontier developers to provide anonymous internal disclosure mechanisms through which covered employees may report concerns, with monthly status updates on the developer's investigation of disclosures (Section 1107.1(e)(1)). These whistleblower protections apply to employees ``responsible for assessing, managing, or addressing risk of critical safety incidents'' (Section 1107(b)). RAISE does not include equivalent whistleblower provisions.

These mechanisms are valuable but structurally limited. Incident reporting captures failures but not the ongoing operations that precede them. Whistleblower protections, where they exist, depend on insiders recognizing issues, understanding their regulatory significance, and accepting career risks despite legal protections. Neither mechanism provides regulators with visibility into normal internal operations---the baseline against which to identify concerning deviations. Regulators may learn of the eventual accident but remain blind to the trajectory that produced it.

\textbf{Proactive disclosure addresses timing but has verification challenges.} Some frameworks mandate affirmative disclosure rather than waiting for incidents or whistleblowers. California's SB 53 and New York's RAISE Act both require large frontier developers to submit quarterly confidential summaries of catastrophic risk assessments resulting from internal use (SB 53 Section 22757.12(d); RAISE Section 1422(b)(1)).

However, proactive disclosure creates verification challenges. Both statutes explicitly mandate ``summaries'' rather than underlying evaluation data, methodological details, or raw incident logs. Regulators must accept company characterizations at face value without independent technical mechanisms to verify claims about system capabilities or operations. Both frameworks permit transmission of reports to other agencies: under SB 53, the Attorney General or Office of Emergency Services may transmit reports to the Legislature, Governor, federal government, or appropriate state agencies (Section 22757.13(e)(1)); under RAISE, the Office may transmit reports to other governmental entities including the Attorney General (Section 1422(e)). Yet no verification mechanism exists in either framework to assess whether submitted characterizations accurately reflect internal operations.

The EU framework exhibits a parallel limitation: for internal-only systems, regulators primarily receive the Safety and Security Framework (Safety Commitment 1) defined as a ``high-level description'' of processes, rather than the detailed Safety and Security Model Reports (Safety and Security Commitment 7) that contain granular evaluation results and capability descriptions. Those detailed reports are explicitly conditioned on ``placing a model on the market.'' European regulators may see corporate policy and eventual incidents, but remain blind to the specific capabilities of internal models operating between those extremes.

\textbf{Regulators can compel detailed information, but only if they already know what to request and from whom.} Several frameworks grant regulators authority to compel detailed information, but only upon request. AIREA empowers the Department of Energy to obtain ``code, training data, and model weights'' for evaluation. The EU's framework includes ``upon request'' authorities for additional documentation. These provisions could theoretically provide deep visibility into internal systems. Yet on-request authority presupposes that regulators know what to request and from whom. Without affirmative disclosure obligations that surface the existence of internal systems meeting capability thresholds, regulators lack the informational predicate to exercise compulsory authority. RAISE's registration requirement (Section 1428) helps identify \textit{who} is covered but does not reveal \textit{what} internal systems those developers operate or how capable those systems are. The mechanisms exist but remain dormant absent external signals that internal-only systems may never generate. On-request authority addresses the depth of information regulators can obtain but not the antecedent problem of identifying systems warranting scrutiny.

\textbf{Common challenges: a lack of comprehensive disclosure mechanisms that provide both the breadth and depth needed for meaningful oversight.} Each regulatory approach to information asymmetry addresses part of the problem while leaving structural gaps. Reactive mechanisms capture failures but not ongoing operations. Proactive disclosure provides regular visibility but at summary depth that resists verification. Registration requirements help identify covered entities but not the specific systems they operate internally. On-request authority enables deep access but requires prior knowledge of what to request. The underlying challenge is that internal deployment inverts the normal information environment: external deployment creates observable market signals that trigger regulatory attention, while internal deployment generates value precisely through invisibility to external observers. Regulations designed to process information flowing from market activity struggle when the most consequential systems never enter markets at all.

\section{Discussion} \label{sec:discussion}

The three gaps we identified---scope ambiguities, static compliance criteria, and information asymmetries---appear across different regulations and jurisdictions. This pattern suggests a persistent challenge in applying traditional regulatory paradigms to AI systems deployed internally. Understanding why these gaps persist, and what approaches might address them, requires examining both the difficulties behind them and the range of potential responses.

\subsection{Why Do These Gaps Persist?} \label{sec:persist}

\textbf{The boundaries between development, testing, and deployment resist clean regulatory line-drawing when systems serve dual purposes.} For internal deployment, the boundaries between traditional regulatory categories become ambiguous, resisting clear rules. Using an AI system to automate portions of research, optimize business operations, or analyze user behavior is simultaneously operational use (the system performs its intended function, generating value) and potentially part of ongoing development and testing (refining capabilities, gathering data for improvements). Drawing the line too narrowly risks imposing unintentional compliance burdens on legitimate research activities and potentially chilling beneficial experimentation. Drawing it too broadly creates gameable loopholes discussed in Section 3.

\textbf{Regulators' access to detailed information creates tensions with legitimate confidentiality interests.} Effective oversight of internal deployment requires information: internal use cases, evaluation results, and incident reports. But frontier AI development involves legitimate trade secrets: models, how they are developed, and how they are evaluated represent significant R\&D investments. Companies have legitimate concerns that forced disclosure risks leaks to competitors or public disclosure through transparency laws or security breaches. Solutions like trusted intermediaries (e.g., DOE evaluation programs) are complex to establish, requiring secure facilities, personnel with appropriate expertise, and protocols that protect commercial secrets while enabling meaningful evaluation. Yet other high-stakes domains have solved analogous challenges with confidentiality protections that prevent public disclosure while enabling regulatory verification \citep{andersonsamways2025regulatoryprecedent, govai2024structuredaccess, shevlane2022structured}. The information asymmetry problem identified in \Cref{sec:asymmetry} is not inevitable. Other high-stakes domains have developed mechanisms that enable regulatory verification while protecting commercial confidentiality. Building such mechanisms requires policymakers to first recognize that internal deployment poses distinct oversight challenges.

\textbf{Market incentive failures compound information asymmetry by undermining the security needed to protect high-value internal systems.} The information asymmetry discussed directly above creates particular concerns for internal deployment because these systems become high-value targets: a model used internally to automate research or optimize operations may be more capable and valuable than anything publicly released. Yet competitive pressures cut against adequate security for these systems. Consider the security measures needed to protect a highly capable internal model: air-gapped development environments isolated from internet-connected systems, access controls limiting who can interact with a model, hardware security modules to prevent exfiltration of parameters, insider threat monitoring, and vetting of the technology stack to prevent supply chain compromise \citep{rand2024securityweights}. Implementing such protections requires significant upfront investment and slows deployment \citep{rand2024securitybrief, grunewald2025datacentersecurity}. Meanwhile, competitive pressures create powerful incentives to deploy quickly: a company that reaches transformative internal capabilities first gains compounding advantages in research productivity or operational efficiency, even if doing so means accepting greater security risks. This dynamic intensifies in a geopolitical context, where state actors may explicitly target internal models as high-value intelligence priorities.

\subsection{Approaches and Tradeoffs} \label{sec:approaches}

Multiple approaches could address internal deployment gaps, each with distinct tradeoffs. Rather than advocating specific solutions, we map the design space to clarify choices and implications.

\textbf{Scope ambiguity could be addressed by triggering obligations based on what systems do rather than how organizations characterize them.} The scope ambiguity identified in \Cref{sec:scope} stems from regulations assuming a clear develop, test, deploy sequence. Internal AI use blurs these boundaries, because a system accelerating development of its successor is simultaneously ``R\&D'' by organizational characterization and ``operational'' by function. Rather than attempting to sharpen the boundary between testing and deployment, regulations could trigger obligations based on functional characteristics: whether a system performs consequential work the organization relies on, accesses sensitive infrastructure, or operates with reduced safeguards. Mechanical thresholds offer an alternative; cumulative inference compute limits, for instance, would be harder to game and would also bound security exposure during pre-compliance operation, though such thresholds may not track risk directly. The core tradeoff is that functional criteria are conceptually well-targeted but depend on self-reporting regulators cannot easily verify, while mechanical thresholds are verifiable but risk being over- or under-inclusive. Implementation could occur through regulatory guidance, revised statutory definitions, or delegated rulemaking that allows criteria to evolve as practices mature.

\textbf{Regulations could establish assessment triggers tied to capability indicators, use cases, or incident patterns rather than just fixed periodic reviews.} Beyond fixed periodic reviews, regulations could establish triggers for assessment based on capability indicators, use cases, or incident patterns. Triggers might include demonstrated capabilities based on evaluations (e.g., sophisticated deception, autonomous operation beyond specified bounds, advanced capabilities in biological or chemical domains), usage for pre-specified types of tasks, integration into new high-stakes applications, or patterns of concerning behaviors reported through incident systems. This addresses both the static compliance gap (by creating assessment points tied to actual system evolution) and enables more targeted oversight than blanket periodic requirements. Tradeoffs include administrative complexity, potential for gaming (companies training to pass evaluations), challenges with updating criteria as capabilities evolve, and determining appropriate threshold levels that balance sensitivity with practicality.

\textbf{Ongoing oversight mechanisms could replace point-in-time assessment for systems that evolve continuously.} Rather than point-in-time pre-deployment assessment, regulations could require ongoing oversight scaled to system capability and deployment context. In practice, this might range from continuous logging requirements (companies maintain auditable records of system modifications, capability evaluations, and incident reports) to periodic regulator access for inspection. Automated monitoring tools could supplement human oversight, running in parallel with system operation. This directly addresses the static compliance gap and is particularly suited to internal deployment, where systems may evolve without the release cycles that create natural checkpoints. The primary tradeoffs are compliance burden, which scales with oversight intensity, and the need for regulators or auditors with sufficient technical expertise to interpret what they observe.

\textbf{Comprehensive reporting frameworks could enable meaningful oversight while maintaining confidentiality protections.} Comprehensive reporting frameworks could allow for meaningful oversight while maintaining confidentiality, potentially by building on California's SB 53 model of confidential quarterly risk assessments. Enhancements might include requiring underlying evaluation data and methodological details, establishing criteria for adequate internal risk assessment, creating escalation procedures when certain capabilities or risk levels are reported, and developing government capacity to assess systems. This addresses information asymmetry while protecting commercial secrets. Implementation requires secure infrastructure and personnel with expertise---substantial but not unprecedented administrative requirements, with precedent in the banking, nuclear, and pharmaceutical sectors \citep{andersonsamways2025regulatoryprecedent}.


\textbf{Credible consequences for non-compliance are necessary for other mechanisms to be effective.} The approaches above focus on generating information and triggering assessments, but their effectiveness depends on credible consequences for non-compliance. Enforcement options include financial penalties scaled to company revenue, restrictions on external deployment until internal compliance is demonstrated, or personal liability for executives who knowingly misrepresent internal system capabilities or uses. The challenge is in detection, as enforcement requires knowing that violations occurred, which returns to the information asymmetry problem. This makes whistleblower protections essential as internal observers may be the only reliable source of information about violations. Mandatory requirements with enforcement and voluntary incentive programs are complementary; incentives encourage disclosure beyond minimums, while penalties ensure minimums are met.

\textbf{Extending existing frameworks to internal deployment need not require regulatory reinvention.} For AI systems already subject to obligations when deployed externally, the core argument for internal deployment oversight is straightforward: internal deployment of comparable systems should trigger the same security and safety requirements that developers already implement for external offerings. Oversight of internal deployment is not about constraining innovation or creating new compliance burdens, but about ensuring that risk management practices already deemed necessary for external systems are not bypassed simply because deployment occurs behind company walls.

\section{Conclusion} \label{sec:conclusion}

Frontier AI regulations share a common structural challenge: they provide limited oversight of internal deployment. We have documented three specific gaps (unclear regulatory scope for operational internal use, static compliance requirements mismatched to continuously evolving systems, and information asymmetries that prevent public oversight) which appear across the EU AI Act and GPAI COP, California's SB 53, New York's RAISE Act, and the US AIREA proposal. These gaps reflect design assumptions from product safety regulation: that development and deployment are distinct phases, that risks emerge primarily from external use, and that oversight attaches at market entry. When AI systems serve dual purposes as both operational tools and development platforms, when high-stakes applications include optimizing user engagement at scale or automating research with access to proprietary systems, or when risks may concentrate during internal operation before any external release, these assumptions create systematic blind spots.

The economic incentives are clear: companies that use AI to perform work currently requiring rare and expensive human talent, gain significant competitive advantages, and will structure their activities to preserve those advantages under regulatory constraints. If regulatory frameworks concentrate oversight on external deployment while providing limited visibility into internal use, the most capable systems may remain internal precisely where oversight mechanisms are least developed. Current frameworks have the appearance of comprehensive AI governance, but provide limited coverage of internal operational deployment. Whether this reflects appropriate prioritization or an unintended consequence of regulatory design assumptions should be determined through careful and explicit analysis, not incidental inheritance from regulatory paradigms designed for products that do not evolve through their own application.

\newpage

\section*{Acknowledgments}
We are grateful to Matteo Pistillo, Oscar Delaney, Ashwin Acharya, Megan Cansfield, Elias Groll, Martin Fukui, and Jake Steckler for discussion and feedback.

\bibliographystyle{apalike}
\bibliography{references}

@misc{anthropic2025claudecode,
  author = {{Anthropic}},
  title = {How {Anthropic} Teams Use {Claude Code}},
  howpublished = {Anthropic Blog},
  year = {2025},
  url = {https://claude.com/blog/how-anthropic-teams-use-claude-code},
}

@misc{openai2025safetyeval,
  author = {{OpenAI}},
  title = {Findings from a pilot Anthropic–OpenAI alignment evaluation exercise: OpenAI Safety Tests},
  howpublished = {OpenAI Blog},
  year = {2025},
  url = {https://openai.com/index/openai-anthropic-safety-evaluation/},
}

@misc{databricks2024guardrails,
  author = {{Databricks}},
  title = {Implementing {LLM} Guardrails for Safe and Responsible Generative {AI} Deployment},
  howpublished = {Databricks Blog},
  year = {2024},
  url = {https://www.databricks.com/blog/implementing-llm-guardrails-safe-and-responsible-generative-ai-deployment-databricks},
}

@article{axios2025amodei,
  author = {Ben Berkowitz},
  title = {Exclusive: Anthropic's Claude is getting better at building itself, Amodei says},
  journal = {Axios},
  year = {2025},
  month = {September},
  day = {17},
  url = {https://www.axios.com/2025/09/17/ai-anthropic-amodei-claude},
  note = {Interview with Dario Amodei discussing Claude's role in AI development}
}

@misc{musk2025tweet,
  author = {Musk, Elon},
  title = {Tweet on X},
  year = {2025},
  howpublished = {Twitter},
  note = {\url{https://x.com/elonmusk/status/1979217645854511402}},
}

@misc{altman2025tweet,
  author = {Altman, Sam},
  title = {Post on {X} regarding automated {AI} research goals},
  howpublished = {X (formerly Twitter)},
  year = {2025},
  url = {https://x.com/sama/status/1983584366547829073},
}

@techreport{rand2024aiautomation,
  author = {Gaurav Sett},
  title = {How {AI} Can Automate {AI} Research and Development},
  institution = {RAND Corporation},
  year = {2024},
  month = {October},
  type = {Commentary},
  url = {https://www.rand.org/pubs/commentary/2024/10/how-ai-can-automate-ai-research-and-development.html}
}

@misc{amodei2025dreamforce,
  author = {Salesforce},
  title = {A Conversation with Dario Amodei and Marc Benioff | Dreamforce 2025},
  year = {2025},
  url = {https://www.youtube.com/watch?v=wUOjTR1511M},
  note = {Accessed: 2025-12-13}
}

@techreport{rand2024securityweights,
  author = {Nevo, Sella and Lahav, Dan and Karpur, Ajay and Bar-On, Yogev and Bradly, Henry Alexander and Alstott, Jeff},
  title = {Securing {AI} Model Weights: Preventing Theft and Misuse of Frontier Models},
  institution = {RAND Corporation},
  year = {2024},
  type = {Research Report},
  url = {https://www.rand.org/pubs/research_reports/RRA2849-1.html}
}

@techreport{rand2024securitybrief,
  author = {Nevo, Sella and Lahav, Dan and Karpur, Ajay and Bar-On, Yogev and  Alexander Bradly, Henry and Alstott, Jeff},
  title = {A Playbook for Securing AI Model Weights},
  institution = {RAND Corporation},
  year = {2024},
  type = {Research Summary},
  url = {https://www.rand.org/pubs/research_briefs/RBA2849-1.html}
}

@article{casper2025pitfalls,
  author = {Casper, Stephen and Krueger, David and Hadfield-Menell, Dylan},
  title = {Pitfalls of Evidence-Based {AI} Policy},
  journal = {arXiv preprint arXiv:2502.09618},
  year = {2025},
  url = {https://arxiv.org/abs/2502.09618}
}

@article{pistillo2025internal,
    author ={Pistillo, Matteo} ,
    title = {Internal Deployment in the EU AI Act},
    journal = {arXiv preprint arXiv:2512.05742},
    year = {2025},
    url = {https://arxiv.org/abs/2512.05742}
}

@article{shevlane2022structured,
  author = {Shevlane, Toby},
  title = {Structured Access: An Emerging Paradigm for Safe {AI} Deployment},
  journal = {arXiv preprint arXiv:2201.05159},
  year = {2022},
  url = {https://arxiv.org/abs/2201.05159}
}

@article{stix2025aibehindcloseddoors,
  author = {Stix, Charlotte and Pistillo, Matteo and Sastry, Girish and Hobbhahn, Marius and Ortega, Alejandro and Balesni, Mikita and Hallensleben, Annika and Goldowsky-Dill, Nix and Sharkey, Lee},
  title = {{AI} Behind Closed Doors: A Primer on The Governance of Internal Deployment},
  journal = {arXiv preprint arXiv:2504.12170},
  year = {2025},
  url = {https://arxiv.org/abs/2504.12170}
}

@techreport{acharya2025internalrisks,
  author = {Acharya, Ashwin and Delaney, Oscar},
  title = {Managing Risks from Internal {AI} Systems},
  institution = {Institute for AI Policy and Strategy (IAPS)},
  year = {2025},
  url = {https://www.iaps.ai/research/managing-risks-from-internal-ai-systems}
}

@techreport{andersonsamways2025regulatoryprecedent,
  author = {Anderson-Samways, Bill},
  title = {AI-Relevant Regulatory Precedents: A Systematic Search Across All Federal Agencies},
  institution = {Institute for AI Policy and Strategy (IAPS)},
  year = {2025},
  url = {https://www.iaps.ai/research/ai-relevant-regulatory-precedent}
}

@techreport{grunewald2025datacentersecurity,
  author = {Grunewald, Erich},
  title = {Accelerating {AI} Data Center Security},
  institution = {Institute for AI Policy and Strategy (IAPS)},
  year = {2025},
  url = {https://www.iaps.ai/research/accelerating-ai-data-center-security}
}

@techreport{govai2024structuredaccess,
  author = {Bucknall, Benjamin S. and Trager, Robert F.},
  title = {Structured Access for Third-Party Research on Frontier {AI} Models},
  institution = {Centre for the Governance of AI},
  year = {2024},
  url = {https://www.governance.ai/research-paper/structured-access-for-third-party-research-on-frontier-ai-models}
}

@article{bommasani2025advancing,
  author = {Bommasani, Rishi and Arora, Sanjeev and Chayes, Jennifer and Choi, Yejin and Cu{\'e}llar, Mariano-Florentino and Fei-Fei, Li and Ho, Daniel E. and Jurafsky, Dan and Koyejo, Sanmi and Lakkaraju, Hima and Narayanan, Arvind and Nelson, Alondra and Pierson, Emma and Pineau, Joelle and Singer, Scott and Varoquaux, Ga{\"e}l and Venkatasubramanian, Suresh and Stoica, Ion and Liang, Percy and Song, Dawn},
  title = {Advancing Science- and Evidence-Based {AI} Policy},
  journal = {Science},
  year = {2025},
  doi = {10.1126/science.adu8449},
  url = {https://www.science.org/doi/10.1126/science.adu8449}
}

@misc{euaiact2024,
  author = {{European Union}},
  title = {Artificial Intelligence Act ({EU} {AI} Act)},
  year = {2024},
  howpublished = {Official Journal of the European Union, Regulation (EU) 2024/1689},
  url = {https://eur-lex.europa.eu/eli/reg/2024/1689/oj}
}

@misc{casb532025,
  author = {{California State Legislature}},
  title = {Senate Bill 53: Transparency in Frontier Artificial Intelligence Act},
  year = {2025},
  note = {Chapter 138, Statutes of 2025. Authored by Sen. Scott Wiener},
  url = {https://leginfo.legislature.ca.gov/faces/billTextClient.xhtml?bill_id=202520260SB53}
}

@misc{nyraiseact2025,
  author = {{New York State Legislature}},
  title = {Assembly Bill A6453-A: Responsible {AI} Safety and Education Act ({RAISE} Act)},
  year = {2025},
  url = {https://legislation.nysenate.gov/pdf/bills/2025/A6453A}
}

@misc{hawley2025airiskevaluation,
  author = {Hawley, Josh and Blumenthal, Richard},
  title = {Artificial Intelligence Risk Evaluation Act of 2025},
  year = {2025},
  howpublished = {119th United States Congress, Senate Bill S.2938},
  note = {Introduced September 29, 2025. Referred to Committee on Commerce, Science, and Transportation},
  url = {https://www.congress.gov/bill/119th-congress/senate-bill/2938/text}
}

@misc{nrc2020oversight,
  author = {{U.S. Nuclear Regulatory Commission}},
  title = {How We Regulate: Inspection and Safety Oversight},
  year = {2020},
  url = {https://www.nrc.gov/about-nrc/regulatory/safety-oversight},
  note = {Last reviewed/updated July 7, 2020}
}

@misc{fda2015indprotocols,
  author = {{U.S. Food and Drug Administration}},
  title = {{IND} Applications for Clinical Investigations: Clinical Protocols},
  year = {2015},
  url = {https://www.fda.gov/drugs/investigational-new-drug-ind-application/ind-applications-clinical-investigations-clinical-protocols},
  note = {Content current as of October 9, 2015}
}

\end{document}